\documentclass{article}

\usepackage{iclr2026_conference,times}

\usepackage{amsmath,amsfonts,bm}

\def\eqref#1{equation~\ref{#1}}

\def\1{\bm{1}}

\DeclareMathAlphabet{\mathsfit}{\encodingdefault}{\sfdefault}{m}{sl}
\SetMathAlphabet{\mathsfit}{bold}{\encodingdefault}{\sfdefault}{bx}{n}

\usepackage{hyperref}
\usepackage{url}
\usepackage{booktabs}
\usepackage{multirow}
\usepackage{graphicx}
\usepackage{amsmath,amssymb}
\usepackage{enumitem}
\usepackage{array}
\usepackage{xcolor}
\usepackage{microtype}
\usepackage{placeins}
\usepackage{float}
\hypersetup{hidelinks}

\iclrfinalcopy

\newcommand{\method}{\textsc{PAVE}}
\newcommand{\trace}{\mathcal{T}}

\title{PAVE: Premise-Grounded Answer Validation and Editing for Retrieval-Augmented LLMs}

\author{Tianyi Huang\thanks{Corresponding author.} \\
Ryquo \\
\texttt{tianyi@ryquo.com}
\And
Caden Yang \\
App-In Club \\
\texttt{caden@appinclub.org}
\And
Emily Yin \\
App-In Club \\
\texttt{emily@appinclub.org}
\And
Eric Wang \\
App-In Club \\
\texttt{eric.wang@appinclub.org}
\And
Michael Zhang \\
App-In Club \\
\texttt{michael@appinclub.org}
}

\begin{document}
\maketitle

\begin{abstract}
Retrieval-augmented language models can retrieve relevant evidence yet still commit to answers before explicitly checking whether the retrieved context supports the conclusion. We present \method\ (\emph{Premise-Grounded Answer Validation and Editing}), an inference-time validation layer for evidence-grounded question answering. \method\ decomposes retrieved context into question-conditioned atomic facts, drafts an answer, scores how well that draft is supported by the extracted premises, and revises low-support outputs before finalization. The resulting trace makes answer commitment auditable at the level of explicit premises, support scores, and revision decisions. In controlled ablations with a fixed retriever and backbone, \method\ outperforms simpler post-retrieval baselines in two evidence-grounded QA settings, with the largest gain reaching 32.7 accuracy points on a span-grounded benchmark. We view these findings as proof-of-concept evidence that explicit premise extraction plus support-gated revision can strengthen evidence-grounded consistency in retrieval-augmented LLM systems.
\end{abstract}

\section{Introduction}

Retrieval-augmented language models are increasingly used in settings where answers must be not only fluent, but also well supported by the evidence shown to the model. In medicine, science, law, and policy, a response can be harmful even when it sounds plausible if its conclusion is unsupported by the retrieved premises or if it overlooks an exception stated in the evidence. Recent work shows that LLMs remain brittle under hallucination, conflicting evidence, and multi-step reasoning demands, including in retrieval-augmented settings intended to improve grounding \citep{ji2023survey, tonmoy2024survey, wang2023survey, feldman2024ragged}.

Retrieval-augmented generation (RAG) addresses part of this problem by conditioning answer generation on external documents rather than solely on parametric memory \citep{lewis2021rag}. Yet standard RAG still leaves a consequential gap between \emph{retrieving evidence} and \emph{verifying that a conclusion is supported by that evidence}. In many pipelines, the model consumes one or more passages and emits a final answer in a single step. When that answer is wrong, users often cannot tell whether the failure came from retrieval, from overlooking a key sentence, from contradictory premises, or from unsupported inference beyond the evidence.

We study this gap at a deliberately narrow but practically important level: premise-grounded consistency between retrieved evidence and generated answers. The question is not whether the model can produce a longer free-form explanation, but whether its final answer remains faithful to an explicit set of evidence units derived from the retrieved context. This target is narrower than full formal reasoning, but it captures a common failure mode in deployed RAG systems: the model should have revised, softened, or withheld a conclusion once the supporting premises were inspected more carefully.

\begin{figure}[t]
    \centering
    \includegraphics[width=0.98\linewidth]{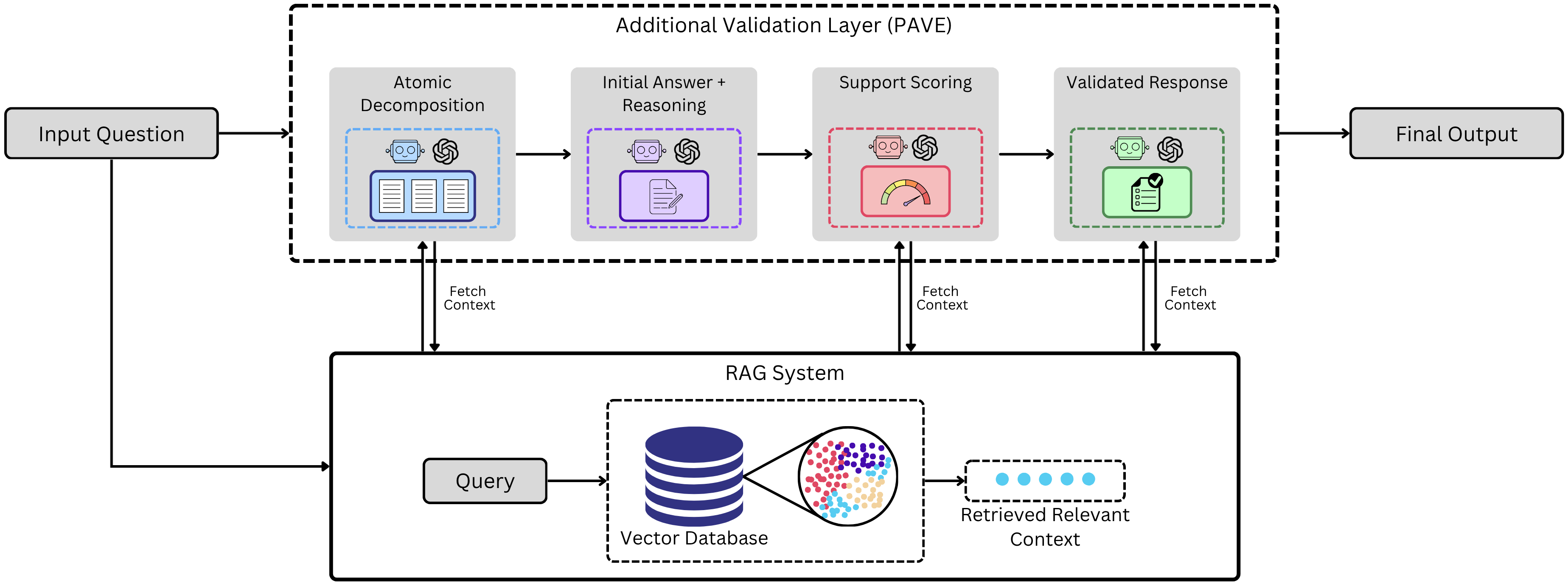}
    \caption{High-level view of \method. Instead of committing immediately to a final answer, the system inserts an explicit validation stage between retrieval and answer finalization: retrieved context is decomposed into atomic facts, used to draft an answer and short rationale, scored for evidence support, and revised only when the support signal is insufficient.}
    \label{fig:intro-overview}
\end{figure}

Recent research suggests two complementary directions for addressing this issue. First, fine-grained retrieval work shows that smaller evidence units, adaptive chunking, and query refinement can improve downstream question answering by matching retrieval granularity more closely to the task \citep{chen2024densex, ma2023rewrite, chan2024rqrag, asai2024selfrag, yan2024crag, jeong2024adaptiverag, sarthi2024raptor}. Second, verification-oriented methods such as SelfCheckGPT and Chain-of-Verification show that inference-time checking can reduce unsupported generations \citep{manakul2023selfcheckgpt, dhuliawala2024cove}. Our proposal combines these perspectives in a lightweight post-retrieval layer.

We present \textbf{\method} (\emph{Premise-Grounded Answer Validation and Editing}), a modular workflow that converts retrieved context into atomic facts, drafts an answer, scores its premise support, and revises low-support answers before returning the final response. The method name emphasizes the central intervention: before answer commitment, the model is asked to make its working premises explicit, validate the draft against them, and edit low-support outputs. Our aim is not to present a fully optimized end-to-end RAG stack. Instead, we study \method\ as a controlled validation interface: the retriever and backbone are held fixed so that the contribution of atomic decomposition and support-gated revision can be examined directly. 

\paragraph{Contributions.}
We make three contributions. First, we introduce \method, a post-retrieval validation layer that combines atomic decomposition, support scoring, and thresholded revision to reduce unsupported answer commitment. Second, we evaluate the method as a controlled ablation study on a fixed retriever and backbone, which isolates the effect of the validation layer and positions the paper as a controlled study. Third, we clarify the interpretability claim by formalizing the audit trace and by providing a paired SQuAD diagnostic that shows how \method\ changes decisions on the same examples.

\section{Related Work}

\subsection{Verification and hallucination mitigation}

A large literature studies hallucination detection and mitigation in LLMs \citep{ji2023survey, tonmoy2024survey, wang2023survey}. Some methods rely on self-consistency or self-checking. SelfCheckGPT, for example, measures divergence across multiple sampled generations and treats disagreement as a signal of factual uncertainty \citep{manakul2023selfcheckgpt}. Chain-of-Verification drafts an answer, asks targeted verification questions, answers them independently, and then synthesizes a revised response \citep{dhuliawala2024cove}. These methods show that inference-time verification can improve reliability without retraining the base model.

\method\ shares this inference-time perspective, but differs in its verification target. Rather than checking agreement among multiple generations alone, it checks whether a draft conclusion is supported by an explicit question-conditioned set of evidence units extracted from retrieved context. This makes the validation problem more local and easier to audit: the central question becomes whether the answer is supported by the retrieved premises, not merely whether the model agrees with itself.

\subsection{Fine-grained retrieval and RAG optimization}

The retrieval side of the literature has moved well beyond fixed passage retrieval. Query rewriting methods improve retrieval by transforming user questions into retrieval-friendly forms \citep{ma2023rewrite, chan2024rqrag}. Adaptive systems such as Self-RAG, CRAG, and Adaptive-RAG dynamically decide when to retrieve or how aggressively to critique retrieved evidence \citep{asai2024selfrag, yan2024crag, jeong2024adaptiverag}. Hierarchical retrieval methods such as RAPTOR support long-document reasoning through recursive summaries \citep{sarthi2024raptor}. Other work argues that retrieval granularity itself matters: Dense X Retrieval shows that proposition-level indexing can outperform sentence- or passage-level retrieval for open-domain QA \citep{chen2024densex}.

\method\ is complementary to these directions. It does not prescribe a specific retriever, chunking strategy, or query-rewriting policy. Instead, it operates after retrieval and improves the \emph{reasoning interface} between retrieved context and answer generation. Better retrieval still helps \method, but \method\ addresses a distinct failure mode: unsupported answer commitment even when useful evidence is already present.

\subsection{Logical reasoning and solver-augmented methods}

Work on logical reasoning benchmarks shows that LLMs remain unreliable on formal and semi-formal reasoning tasks despite strong surface fluency. FOLIO evaluates natural-language reasoning with first-order logic structure \citep{han2022folio}, while LogicBench tests a broader range of logical patterns, including negation and non-monotonic reasoning \citep{parmar2024logicbench}. Solver-augmented systems such as Logic-LM combine LLMs with symbolic solvers to improve faithfulness on logic-heavy tasks \citep{pan2023logiclm}. Related work on logical preference consistency argues that robust reasoning should satisfy invariances such as transitivity and negation consistency \citep{liu2024logicconsistency}.

Compared with solver-based pipelines, \method\ does not perform formal proof search. Instead, it introduces an explicit premise-grounded validation step for evidence-based question answering. This makes it useful both as a standalone consistency layer and as a front end to stronger downstream verifiers: the atomic facts produced by \method\ provide a natural intermediate representation that can be checked by an entailment model or symbolic module.

\section{Method}

\begin{figure}[h]
    \centering
    \includegraphics[width=\linewidth]{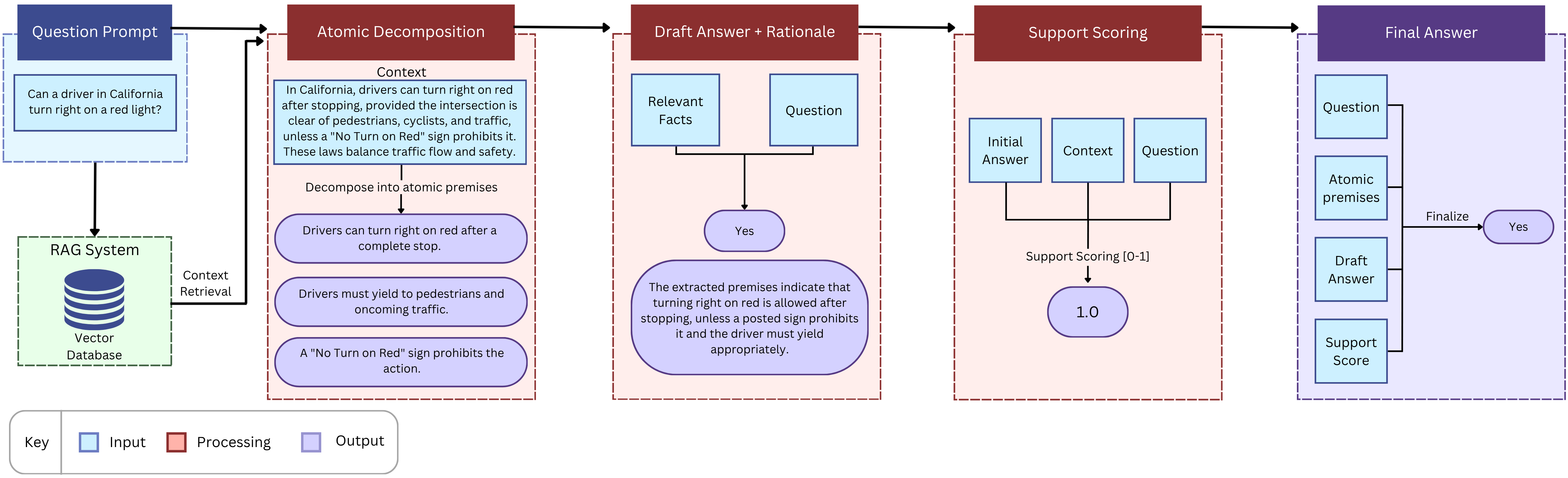}
    \caption{Illustrative execution trace for \method\ on a traffic-rule question. The retrieved context is decomposed into atomic premises, used to produce a draft answer and rationale, scored for support, and then finalized. In this example, the decomposition makes the governing rule, the relevant condition, and the exception explicit before final answer commitment.}
    \label{fig:method-trace}
\end{figure}

\subsection{Problem setup}

Let $q$ denote a user question and let a retriever return context
\begin{equation}
C = \mathcal{R}(q) = \{c_1,\ldots,c_m\},
\end{equation}
where each $c_j$ is a retrieved passage or document chunk. Standard RAG directly maps $(q,C)$ to an answer. \method\ inserts an explicit validation layer that separates evidence extraction, draft generation, support scoring, and answer finalization.

The atomic decomposition stage produces a question-conditioned fact list
\begin{equation}
F = \mathcal{D}(q,C) = [f_1,\ldots,f_n],
\end{equation}
where each $f_i$ is intended to be a concise, standalone statement that is directly relevant to answering $q$. A generator then drafts an answer and short rationale,
\begin{equation}
(a^{(0)}, r^{(0)}) = \mathcal{G}(q,F),
\end{equation}
and a support scorer returns a scalar score
\begin{equation}
s = \mathcal{S}(q,F,a^{(0)},r^{(0)}) \in [0,1],
\end{equation}
where higher values indicate stronger support of the draft answer by the atomic fact set. Finalization is governed by a fixed revision threshold $\tau \in [0,1]$:
\begin{equation}
a^{\star} =
\begin{cases}
a^{(0)}, & \text{if } s \ge \tau, \\
\mathcal{U}(q,F,a^{(0)},r^{(0)},s), & \text{if } s < \tau,
\end{cases}
\end{equation}
where $\mathcal{U}$ is a revision step invoked only for low-support drafts. The corresponding audit trace is
\begin{equation}
\trace = (F, a^{(0)}, r^{(0)}, s, a^{\star}).
\end{equation}
This trace is the main explainability artifact returned by the system.

\subsection{Atomic decomposition as premise extraction}

The first stage converts retrieved context into a compact set of premises. In practice, this decomposition is question-conditioned: the model is asked to retain only information needed to answer $q$, and to express that information in short, self-contained statements rather than long copied passages. This has two benefits. First, it reduces distraction from irrelevant text in later stages. Second, it makes support checking more precise because exceptions and preconditions remain explicit instead of being buried inside longer paragraphs. In qualitative traces, the most useful decompositions separate the main claim, the relevant condition, and any exception into distinct premises, since those are precisely the elements the support scorer must inspect.

The core \method\ pipeline does \emph{not} require learned or hand-crafted importance weights over atomic facts. We explored salience weighting in an earlier prototype, but it proved less stable than the simpler decomposition-plus-scoring design. For clarity, we therefore treat importance weighting as an ablation rather than part of the main method.

\subsection{Support scoring and revision}

Given the atomic fact list $F$, the generator produces a draft answer $a^{(0)}$ and short rationale $r^{(0)}$. A separate scoring prompt then evaluates whether the conclusion is sufficiently supported by the premises. This score is best interpreted as a \emph{premise-support signal}: values near 1 indicate that the answer follows directly or is strongly supported by the extracted facts, while lower values indicate weak support, contradiction, or missing conditions.

The score is not merely reported; it changes system behavior. High-support drafts are finalized immediately, while low-support drafts trigger revision. This gating mechanism is central to the design: the point of the scorer is not only to annotate the answer, but to prevent premature commitment when the available premises do not support the conclusion strongly enough.

\subsection{What the trace adds beyond direct LLM interaction}

A direct interaction with an LLM can already produce an answer and an explanation, so it is reasonable to ask what \method\ adds. The key difference is structural. A free-form explanation is post hoc and unconstrained; it does not tell the user which evidence units were explicitly considered, whether the answer crossed a validation threshold, or whether a revision step was triggered. By contrast, \method\ returns a machine-readable trace with three concrete affordances: evidence locality through atomic premises, an explicit decision signal through the support score, and failure localization through the separation of retrieval, draft generation, and revision. This is why the method is useful even when the final answer text overlaps with what a strong LLM might have produced directly.

\FloatBarrier

\section{Experimental Setup}

\subsection{Datasets}

We report results on two evidence-grounded QA settings.

\paragraph{PubMedQA.}
PubMedQA is a biomedical benchmark whose answers are restricted to \texttt{yes}, \texttt{no}, or \texttt{maybe} \citep{jin2019pubmedqa}. The \texttt{maybe} label makes the task especially relevant to evidence-sensitive reasoning because inconclusive or mixed evidence must not be forced into an overconfident binary decision. We report results on a 1{,}000-question labeled subset.

\paragraph{SQuAD.}
SQuAD is a factoid QA benchmark in which answers are short spans supported by the context \citep{rajpurkar2016squad}. Relative to PubMedQA, it places greater emphasis on precise evidence alignment and less on three-way uncertainty calibration. We report results on a 1{,}000-example subset and an additional 5{,}000-example subset used as a scale check for the final \method\ pipeline.

\subsection{Ablation design}

The original draft presented several pipelines side by side in a way that obscured the main contribution. We recast them here as a clean ablation study.
\begin{itemize}[leftmargin=*,topsep=2pt,itemsep=1pt]
    \item \textbf{Baseline RAG}: one-shot answer generation from retrieved context with no atomic decomposition and no support scoring.
    \item \textbf{RAG + importance weighting}: an exploratory variant that decomposes context into atomic statements and attaches salience scores, but does not use the final \method\ design.
    \item \textbf{RAG + support scoring}: a two-stage variant that scores a draft answer against retrieved context but does not use atomic decomposition.
    \item \textbf{\method}: the full method, combining atomic decomposition with support scoring and thresholded revision.
\end{itemize}
This organization isolates the two core questions behind the method: whether decomposition helps on its own, whether scoring helps on its own, and whether the combination is stronger than either component in isolation.

\subsection{Implementation notes}

All compared variants use the same underlying LLM backbone, GPT-4o mini, in the original experiments \citep{openai2024gpt4omini}. The retrieval backend is held fixed across the ablations so that observed differences reflect the validation layer rather than a change in retriever or base model. Each stage is implemented with a specialized prompt for decomposition, draft generation, support scoring, and, when needed, revision. Because \method\ is training-free, it introduces no task-specific training cost; its main practical cost is additional inference latency. The full pipeline uses three LLM calls by default and invokes a fourth call only when revision is triggered.

We report answer-level accuracy as the primary metric. For PubMedQA, this is exact agreement with the gold \texttt{yes}/\texttt{no}/\texttt{maybe} label. For SQuAD, the available experiment logs contain binary correctness labels for each prediction. These labels credit semantically correct span-equivalent answers rather than only verbatim string matches, so the reported SQuAD numbers should be read as answer-level accuracy under a fixed shared protocol rather than as canonical EM/F1. All compared SQuAD variants use the same labeling rule.

\section{Results}

\begin{table}[H]
\caption{Main results on two evidence-grounded QA tasks, PubMedQA and SQuAD (1,000-example subset).}
\label{tab:main-results}
\centering
\small
\begin{tabular}{lcc}
\toprule
\textbf{Pipeline} & \textbf{PubMedQA} & \textbf{SQuAD} \\
\midrule
Baseline RAG & 71.20 & 62.40 \\
RAG + importance weighting & 69.80 & 47.70 \\
RAG + support scoring & 71.90 & 63.60 \\
\method\ (full) & \textbf{73.30} & \textbf{95.10} \\
\bottomrule
\end{tabular}
\end{table}

The full \method\ pipeline achieves the strongest performance on both datasets. On PubMedQA, the gain over the baseline is modest but consistent (+2.1 points), corresponding to a 7.3\% relative reduction in error. On SQuAD, the gain is much larger (+32.7 points on the 1{,}000-example subset), which corresponds to an 87.0\% relative reduction in error against the baseline and suggests that atomic premise extraction and support-gated revision are especially effective when the answer must stay tightly aligned with the available evidence.

Support scoring without atomic decomposition yields only a small improvement over baseline, suggesting that a verification stage is useful but limited when it must reason over raw passages. The importance-weighting variant performs worse than the baseline on both datasets, implying that salience annotations can amplify extraction noise when they are treated as part of the decision rule. The strongest results therefore come from the combined design: premise extraction plus support scoring plus revision.

On the larger 5{,}000-example SQuAD subset, the full \method\ pipeline retains 92.76 accuracy. While lower than the 95.10 observed on the smaller subset, this still suggests that the improvement is not confined to a tiny pilot slice. The overall pattern is consistent with the method's design objective: when answers are short and directly recoverable from retrieved evidence, explicitly checking premise support before finalization can substantially reduce avoidable errors.

\subsection{Paired trace analysis on SQuAD}

Logged outputs for both the RAG + support scoring variant and the full \method\ pipeline are available for a 100-example SQuAD slice, which we use for a paired diagnostic comparison. Because this analysis is paired at the example level, it helps explain where the aggregate gain comes from rather than only restating benchmark-level accuracy.

\begin{figure}[t]
    \centering
    \includegraphics[width=0.58\linewidth]{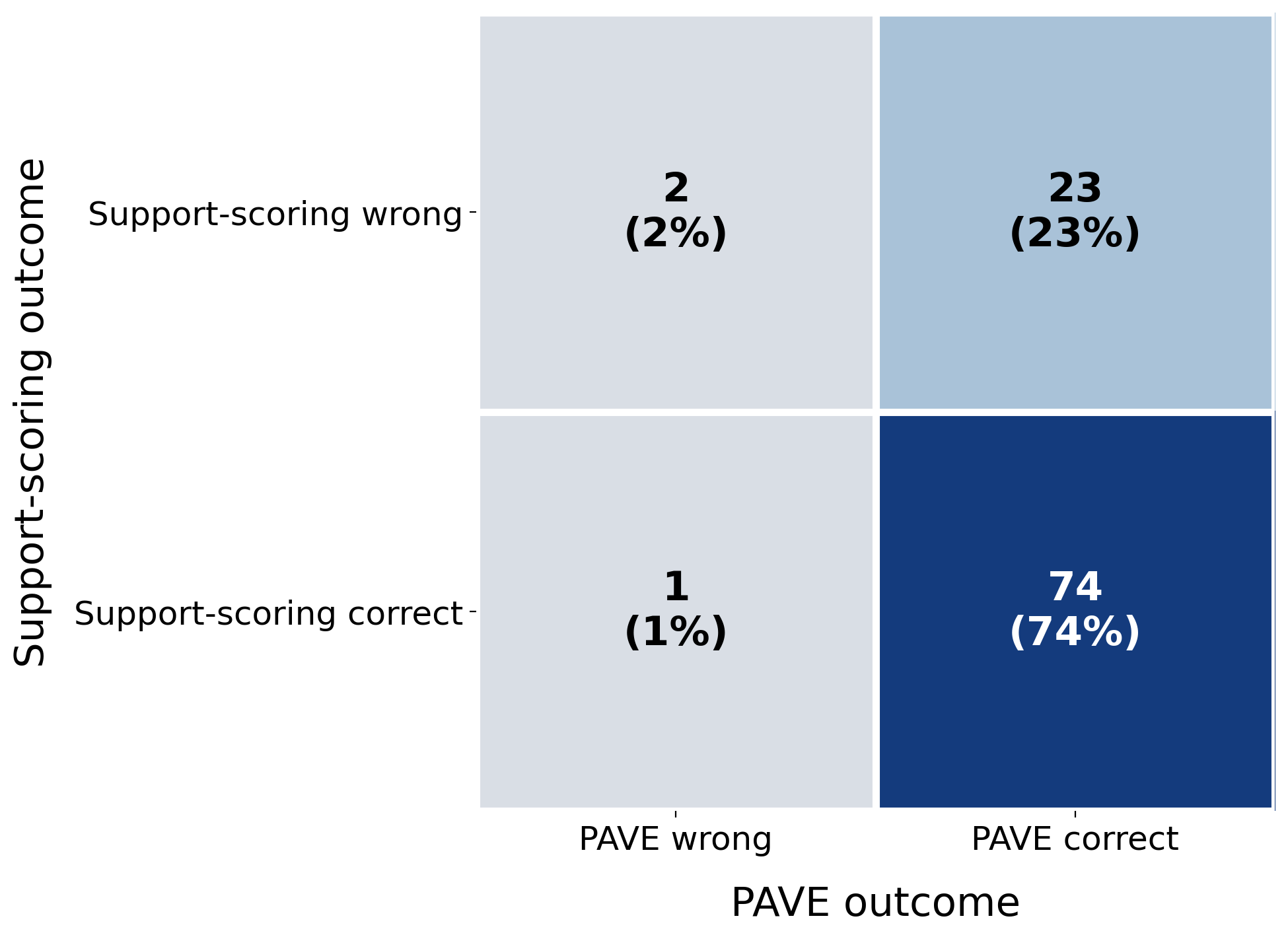}
    \caption{Paired outcome transition on 100 shared SQuAD examples comparing the support-scoring variant with \method.}
    \label{fig:squad-paired}
\end{figure}

Figure~\ref{fig:squad-paired} shows that \method\ corrects 23 of the 25 errors made by the support-scoring variant and introduces only one new error, reducing the error count by 88\% on the same examples. The same slice also shows a marked tightening of answer form: using a simple whitespace token count on the logged outputs, median answer length drops from 71 tokens for the support-scoring variant to 2 tokens for \method, exactly matching the median gold answer length. Within this slice, the largest gain appears on 3--4 token answers, where accuracy rises from 66.7 to 96.7.

Although this evidence is indirect, it is consistent with the claim that decomposition quality matters. When the answer is first grounded in a compact set of question-conditioned premises, the downstream scorer appears better able to localize the answer span and suppress irrelevant explanation. A more direct evaluation of decomposition faithfulness, coverage, and failure modes remains an important next step.

\section{Scope, Limitations, and Future Work}

This paper is best read as a preliminary study of a post-retrieval validation interface. The experiments intentionally keep the retriever and backbone fixed so that the effect of the validation layer can be isolated. That design makes the central claim easier to interpret, but it also narrows the scope of what the current evidence can establish.

Several limitations remain. Paired per-example outputs are currently available only for a diagnostic 100-example SQuAD slice, which constrains how far the present analysis can go beyond aggregate results. In addition, decomposition quality is assessed only indirectly through downstream behavior; a more direct study of decomposition faithfulness, coverage, and failure modes would strengthen the claim that the atomic representation is the right interface for validation. The method also introduces additional inference overhead, since decomposition and support scoring require extra model calls in exchange for greater auditability and control. Finally, because the current experiments hold both the backbone and the retriever fixed, broader evaluation on stronger language models and stronger retrieval pipelines remains an important next step.

From an application perspective, \method\ is most useful when reliability is a main priority. In high-stakes settings, returning an answer together with its supporting premises and a revision signal can make human review substantially easier than relying on a polished but unstructured explanation. A high support score, however, should not be interpreted as a guarantee of truth. The score reflects agreement with the retrieved evidence, and that evidence can itself be incomplete, outdated, or biased.

\section{Conclusion}

This paper studies a simple question with practical consequences for retrieval-augmented language models: should a system commit to an answer before it has explicitly checked whether the retrieved evidence supports that conclusion? \method\ is one answer to that question. By inserting an explicit premise-grounded validation stage between retrieval and final answer commitment, it turns evidence use into something more inspectable and more revisable than one-shot generation alone. Across the reported ablations, the clearest pattern is that support scoring becomes substantially more useful when it operates over compact, question-conditioned premises rather than over raw retrieved passages. We view these results as proof-of-concept evidence for a broader design principle: in evidence-grounded generation, making premises explicit before finalizing an answer can improve not only accuracy, but also the discipline with which answers are formed.

\bibliographystyle{iclr2026_conference}
\bibliography{iclr2026_conference}

@article{ji2023survey,
       title={Survey of Hallucination in Natural Language Generation},
       volume={55},
       ISSN={1557-7341},
       url={http://dx.doi.org/10.1145/3571730},
       DOI={10.1145/3571730},
       number={12},
       journal={ACM Computing Surveys},
       publisher={Association for Computing Machinery (ACM)},
       author={Ji, Ziwei and Lee, Nayeon and Frieske, Rita and Yu, Tiezheng and Su, Dan and Xu, Yan and Ishii, Etsuko and Bang, Ye Jin and Madotto, Andrea and Fung, Pascale},
       year={2023},
       month=mar, pages={1–38} 
}

@misc{tonmoy2024survey,
      title={A Comprehensive Survey of Hallucination Mitigation Techniques in Large Language Models}, 
      author={S. M Towhidul Islam Tonmoy and S M Mehedi Zaman and Vinija Jain and Anku Rani and Vipula Rawte and Aman Chadha and Amitava Das},
      year={2024},
      eprint={2401.01313},
      archivePrefix={arXiv},
      primaryClass={cs.CL},
      url={https://arxiv.org/abs/2401.01313}, 
}

@misc{wang2023survey,
      title={Survey on Factuality in Large Language Models: Knowledge, Retrieval and Domain-Specificity}, 
      author={Cunxiang Wang and Xiaoze Liu and Yuanhao Yue and Xiangru Tang and Tianhang Zhang and Cheng Jiayang and Yunzhi Yao and Wenyang Gao and Xuming Hu and Zehan Qi and Yidong Wang and Linyi Yang and Jindong Wang and Xing Xie and Zheng Zhang and Yue Zhang},
      year={2023},
      eprint={2310.07521},
      archivePrefix={arXiv},
      primaryClass={cs.CL},
      url={https://arxiv.org/abs/2310.07521}, 
}

@misc{lewis2021rag,
      title={Retrieval-Augmented Generation for Knowledge-Intensive NLP Tasks}, 
      author={Patrick Lewis and Ethan Perez and Aleksandra Piktus and Fabio Petroni and Vladimir Karpukhin and Naman Goyal and Heinrich Küttler and Mike Lewis and Wen-tau Yih and Tim Rocktäschel and Sebastian Riedel and Douwe Kiela},
      year={2021},
      eprint={2005.11401},
      archivePrefix={arXiv},
      primaryClass={cs.CL},
      url={https://arxiv.org/abs/2005.11401}, 
}

@misc{feldman2024ragged,
      title={RAGged Edges: The Double-Edged Sword of Retrieval-Augmented Chatbots}, 
      author={Philip Feldman and James R. Foulds and Shimei Pan},
      year={2024},
      eprint={2403.01193},
      archivePrefix={arXiv},
      primaryClass={cs.CL},
      url={https://arxiv.org/abs/2403.01193}, 
}

@misc{manakul2023selfcheckgpt,
      title={SelfCheckGPT: Zero-Resource Black-Box Hallucination Detection for Generative Large Language Models}, 
      author={Potsawee Manakul and Adian Liusie and Mark J. F. Gales},
      year={2023},
      eprint={2303.08896},
      archivePrefix={arXiv},
      primaryClass={cs.CL},
      url={https://arxiv.org/abs/2303.08896}, 
}

@misc{dhuliawala2024cove,
      title={Chain-of-Verification Reduces Hallucination in Large Language Models}, 
      author={Shehzaad Dhuliawala and Mojtaba Komeili and Jing Xu and Roberta Raileanu and Xian Li and Asli Celikyilmaz and Jason Weston},
      year={2023},
      eprint={2309.11495},
      archivePrefix={arXiv},
      primaryClass={cs.CL},
      url={https://arxiv.org/abs/2309.11495}, 
}

@misc{ma2023rewrite,
      title={Query Rewriting for Retrieval-Augmented Large Language Models}, 
      author={Xinbei Ma and Yeyun Gong and Pengcheng He and Hai Zhao and Nan Duan},
      year={2023},
      eprint={2305.14283},
      archivePrefix={arXiv},
      primaryClass={cs.CL},
      url={https://arxiv.org/abs/2305.14283}, 
}

@misc{chan2024rqrag,
      title={RQ-RAG: Learning to Refine Queries for Retrieval Augmented Generation}, 
      author={Chi-Min Chan and Chunpu Xu and Ruibin Yuan and Hongyin Luo and Wei Xue and Yike Guo and Jie Fu},
      year={2024},
      eprint={2404.00610},
      archivePrefix={arXiv},
      primaryClass={cs.CL},
      url={https://arxiv.org/abs/2404.00610}, 
}

@misc{asai2024selfrag,
      title={Self-RAG: Learning to Retrieve, Generate, and Critique through Self-Reflection}, 
      author={Akari Asai and Zeqiu Wu and Yizhong Wang and Avirup Sil and Hannaneh Hajishirzi},
      year={2023},
      eprint={2310.11511},
      archivePrefix={arXiv},
      primaryClass={cs.CL},
      url={https://arxiv.org/abs/2310.11511}, 
}

@misc{yan2024crag,
      title={Corrective Retrieval Augmented Generation}, 
      author={Shi-Qi Yan and Jia-Chen Gu and Yun Zhu and Zhen-Hua Ling},
      year={2024},
      eprint={2401.15884},
      archivePrefix={arXiv},
      primaryClass={cs.CL},
      url={https://arxiv.org/abs/2401.15884}, 
}

@misc{jeong2024adaptiverag,
      title={Adaptive-RAG: Learning to Adapt Retrieval-Augmented Large Language Models through Question Complexity}, 
      author={Soyeong Jeong and Jinheon Baek and Sukmin Cho and Sung Ju Hwang and Jong C. Park},
      year={2024},
      eprint={2403.14403},
      archivePrefix={arXiv},
      primaryClass={cs.CL},
      url={https://arxiv.org/abs/2403.14403}, 
}

@misc{chen2024densex,
      title={Dense X Retrieval: What Retrieval Granularity Should We Use?}, 
      author={Tong Chen and Hongwei Wang and Sihao Chen and Wenhao Yu and Kaixin Ma and Xinran Zhao and Hongming Zhang and Dong Yu},
      year={2024},
      eprint={2312.06648},
      archivePrefix={arXiv},
      primaryClass={cs.CL},
      url={https://arxiv.org/abs/2312.06648}, 
}

@misc{sarthi2024raptor,
      title={RAPTOR: Recursive Abstractive Processing for Tree-Organized Retrieval}, 
      author={Parth Sarthi and Salman Abdullah and Aditi Tuli and Shubh Khanna and Anna Goldie and Christopher D. Manning},
      year={2024},
      eprint={2401.18059},
      archivePrefix={arXiv},
      primaryClass={cs.CL},
      url={https://arxiv.org/abs/2401.18059}, 
}

@misc{han2022folio,
      title={FOLIO: Natural Language Reasoning with First-Order Logic}, 
      author={Simeng Han and Hailey Schoelkopf and Yilun Zhao and Zhenting Qi and Martin Riddell and Wenfei Zhou and James Coady and David Peng and Yujie Qiao and Luke Benson and Lucy Sun and Alex Wardle-Solano and Hannah Szabo and Ekaterina Zubova and Matthew Burtell and Jonathan Fan and Yixin Liu and Brian Wong and Malcolm Sailor and Ansong Ni and Linyong Nan and Jungo Kasai and Tao Yu and Rui Zhang and Alexander R. Fabbri and Wojciech Kryscinski and Semih Yavuz and Ye Liu and Xi Victoria Lin and Shafiq Joty and Yingbo Zhou and Caiming Xiong and Rex Ying and Arman Cohan and Dragomir Radev},
      year={2024},
      eprint={2209.00840},
      archivePrefix={arXiv},
      primaryClass={cs.CL},
      url={https://arxiv.org/abs/2209.00840}, 
}

@misc{parmar2024logicbench,
      title={LogicBench: Towards Systematic Evaluation of Logical Reasoning Ability of Large Language Models}, 
      author={Mihir Parmar and Nisarg Patel and Neeraj Varshney and Mutsumi Nakamura and Man Luo and Santosh Mashetty and Arindam Mitra and Chitta Baral},
      year={2024},
      eprint={2404.15522},
      archivePrefix={arXiv},
      primaryClass={cs.CL},
      url={https://arxiv.org/abs/2404.15522}, 
}

@misc{pan2023logiclm,
      title={Logic-LM: Empowering Large Language Models with Symbolic Solvers for Faithful Logical Reasoning}, 
      author={Liangming Pan and Alon Albalak and Xinyi Wang and William Yang Wang},
      year={2023},
      eprint={2305.12295},
      archivePrefix={arXiv},
      primaryClass={cs.CL},
      url={https://arxiv.org/abs/2305.12295}, 
}

@misc{liu2024logicconsistency,
      title={Aligning with Logic: Measuring, Evaluating and Improving Logical Preference Consistency in Large Language Models}, 
      author={Yinhong Liu and Zhijiang Guo and Tianya Liang and Ehsan Shareghi and Ivan Vulić and Nigel Collier},
      year={2025},
      eprint={2410.02205},
      archivePrefix={arXiv},
      primaryClass={cs.CL},
      url={https://arxiv.org/abs/2410.02205}, 
}

@misc{jin2019pubmedqa,
      title={PubMedQA: A Dataset for Biomedical Research Question Answering}, 
      author={Qiao Jin and Bhuwan Dhingra and Zhengping Liu and William W. Cohen and Xinghua Lu},
      year={2019},
      eprint={1909.06146},
      archivePrefix={arXiv},
      primaryClass={cs.CL},
      url={https://arxiv.org/abs/1909.06146}, 
}

@misc{rajpurkar2016squad,
      title={SQuAD: 100,000+ Questions for Machine Comprehension of Text}, 
      author={Pranav Rajpurkar and Jian Zhang and Konstantin Lopyrev and Percy Liang},
      year={2016},
      eprint={1606.05250},
      archivePrefix={arXiv},
      primaryClass={cs.CL},
      url={https://arxiv.org/abs/1606.05250}, 
}

@misc{openai2024gpt4omini,
      title={GPT-4o mini: advancing cost-efficient intelligence},
      author={{OpenAI}},
      year={2024},
      url={https://openai.com/index/gpt-4o-mini-advancing-cost-efficient-intelligence/}
}

\end{document}